\documentclass[10pt,twocolumn,letterpaper]{article}

\usepackage{cvpr}
\usepackage{lineno}
\usepackage{graphicx}
\usepackage{amsmath}
\usepackage{amssymb}
\usepackage{booktabs}
\usepackage{multirow}
\usepackage{makecell}
\usepackage{ulem}
\usepackage[table]{xcolor}
\usepackage{colortbl}
\newcommand{\thevae}{VA-VAE}
\newcommand{\thedit}{{LightningDiT}}

\definecolor{cvprblue}{rgb}{0.21,0.49,0.74}
\usepackage[pagebackref,breaklinks,colorlinks,allcolors=cvprblue]{hyperref}
\def\confName{CVPR}
\def\confYear{2025}
\title{Reconstruction \textit{vs.} Generation: \\Taming Optimization Dilemma in Latent Diffusion Models}

\author{
Jingfeng Yao$^{1}$, Bin Yang$^{2}$, Xinggang Wang$^{1,\thanks{Corresponding author: xgwang@hust.edu.cn}}$ \\
$^{1}$Huazhong University of Science and Technology \\
$^{2}$Independent Researcher
}

\begin{document}
\maketitle

\begin{abstract}

Latent diffusion models with Transformer architectures excel at generating high-fidelity images. However, recent studies reveal an {\textbf{optimization dilemma}} in this two-stage design: while increasing the per-token feature dimension in visual tokenizers improves reconstruction quality, it requires substantially larger diffusion models and more training iterations to achieve comparable generation performance. 
Consequently, existing systems often settle for sub-optimal solutions, either producing visual artifacts due to information loss within tokenizers or failing to converge fully due to expensive computation costs.
We argue that this dilemma stems from the inherent difficulty in learning unconstrained high-dimensional latent spaces. To address this, we propose aligning the latent space with pre-trained vision foundation models when training the visual tokenizers. Our proposed \textbf{VA-VAE} (Vision foundation model Aligned Variational AutoEncoder) significantly expands the reconstruction-generation frontier of latent diffusion models, enabling faster convergence of Diffusion Transformers (DiT) in high-dimensional latent spaces. 
To exploit the full potential of VA-VAE, we build an enhanced DiT baseline with improved training strategies and architecture designs, termed \textbf{LightningDiT}. 
The integrated system achieves state-of-the-art (SOTA) performance on ImageNet 256$\times$256 generation with an \textbf{FID score of 1.35} while demonstrating remarkable training efficiency by reaching an FID score of 2.11 in just 64 epochs -- representing an over 21$\times$ convergence speedup compared to the original DiT.
Models and codes are available at \url{https://github.com/hustvl/LightningDiT}.

\end{abstract}

\section{Introduction}
\label{sec:intro}

\begin{figure}
    \centering
    \includegraphics[width=\linewidth]{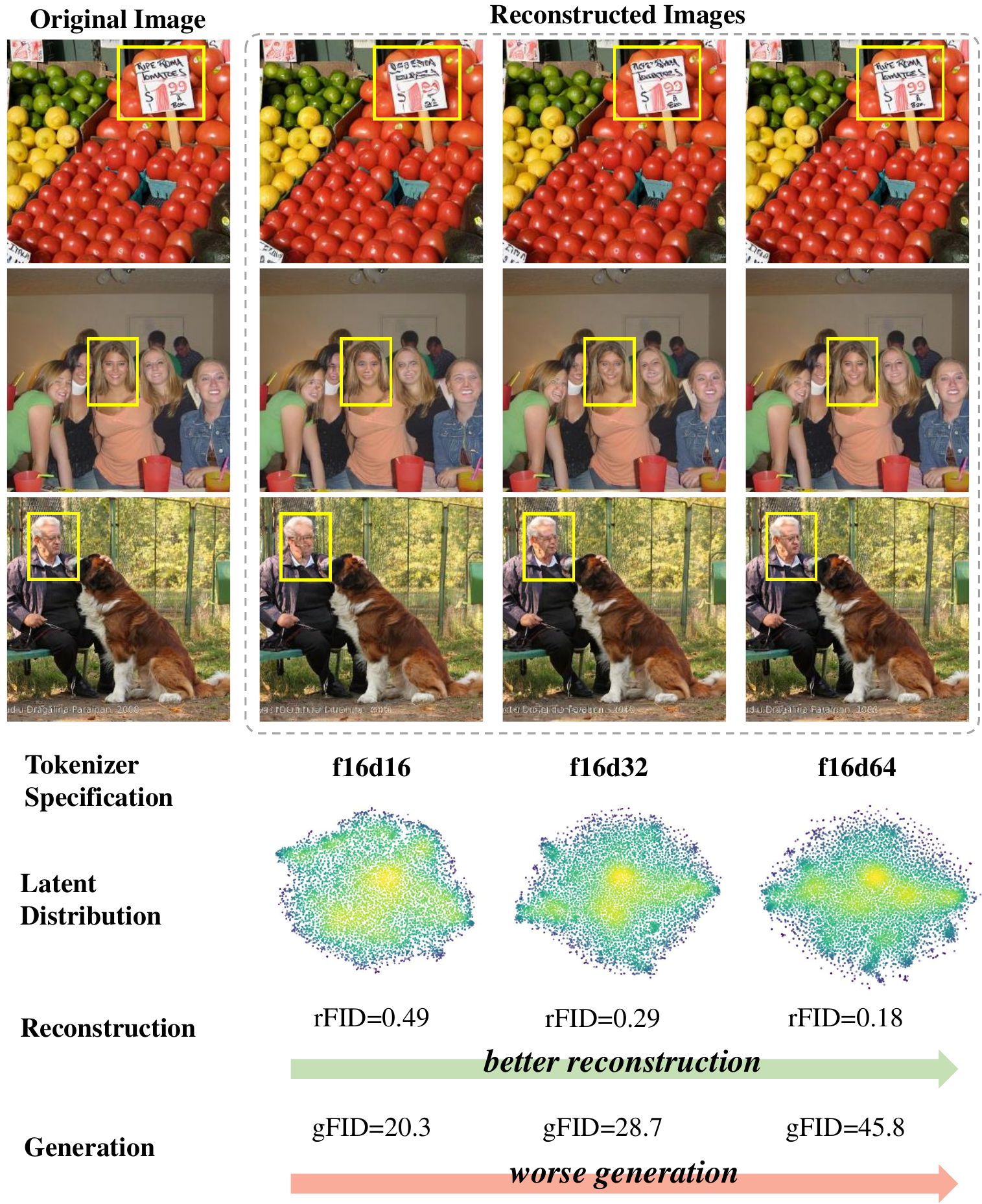}
    \caption{\textbf{Optimization dilemma within latent diffusion models.} In latent diffusion models, increasing the dimension of the visual tokenizer enhances detail reconstruction but significantly reduces generation quality. (In tokenizer specification, ``f" and ``d" represent the downsampling rate and dimension, respectively. All results are evaluated on ImageNet 256$\times$256 dataset with a fixed compute budget during diffusion model training.)}
    \label{fig:optimization_conflict}
\end{figure}

The latent diffusion model~\cite{ldm} utilizes a continuous-valued variational autoencoder (VAE)~\cite{vae}, or visual tokenizer, to compress visual signals and thereby reduce the computational demands of high-resolution image generation. The performance of these visual tokenizers, particularly their compression and reconstruction capabilities, plays a crucial role in determining the overall system effectiveness~\cite{emu, sd3}. A straightforward approach to enhance the reconstruction capability is to increase the feature dimension of visual tokens, which effectively expands the information capacity of the latent representation. Recently, several influential text-to-image works~\cite{sd3, flux, emu} have explored higher-dimensional tokenizers compared to the widely adopted VAE in Stable Diffusion~\cite{sdxl,ldm}, as these tokenizers offer improved detail reconstruction, enabling finer generative quality.

However, as research has advanced, an \textit{optimization dilemma} has emerged between reconstruction and generation performance in latent diffusion models~\cite{sd3,walt,kilian2024computational}. Specifically, while increasing token feature dimension improves the tokenizer's reconstruction accuracy, it significantly degrades the generation performance (see Fig.~\ref{fig:optimization_conflict}). Currently, two common strategies exist to address this issue: the first involves scaling up model parameters, as demonstrated by Stable Diffusion 3~\cite{sd3}, which shows that higher-dimensional tokenizers can achieve stronger generation performance with a significantly larger model capacity—however, this approach requires significantly more training compute, making it prohibitively expensive for most practical applications. The second strategy is to deliberately limit the tokenizer's reconstruction capacity, e.g. Sana~\cite{sana, dcae}, W.A.L.T~\cite{walt}, for faster convergence of diffusion model training. Yet, this compromised reconstruction quality inherently limits the upper bound of generation performance, leading to imperfect visual details in generated results. Both approaches involve inherent trade-offs and do not provide an effective control to the underlying optimization dilemma.

\begin{figure}
    \centering
    \includegraphics[width=0.9\linewidth]{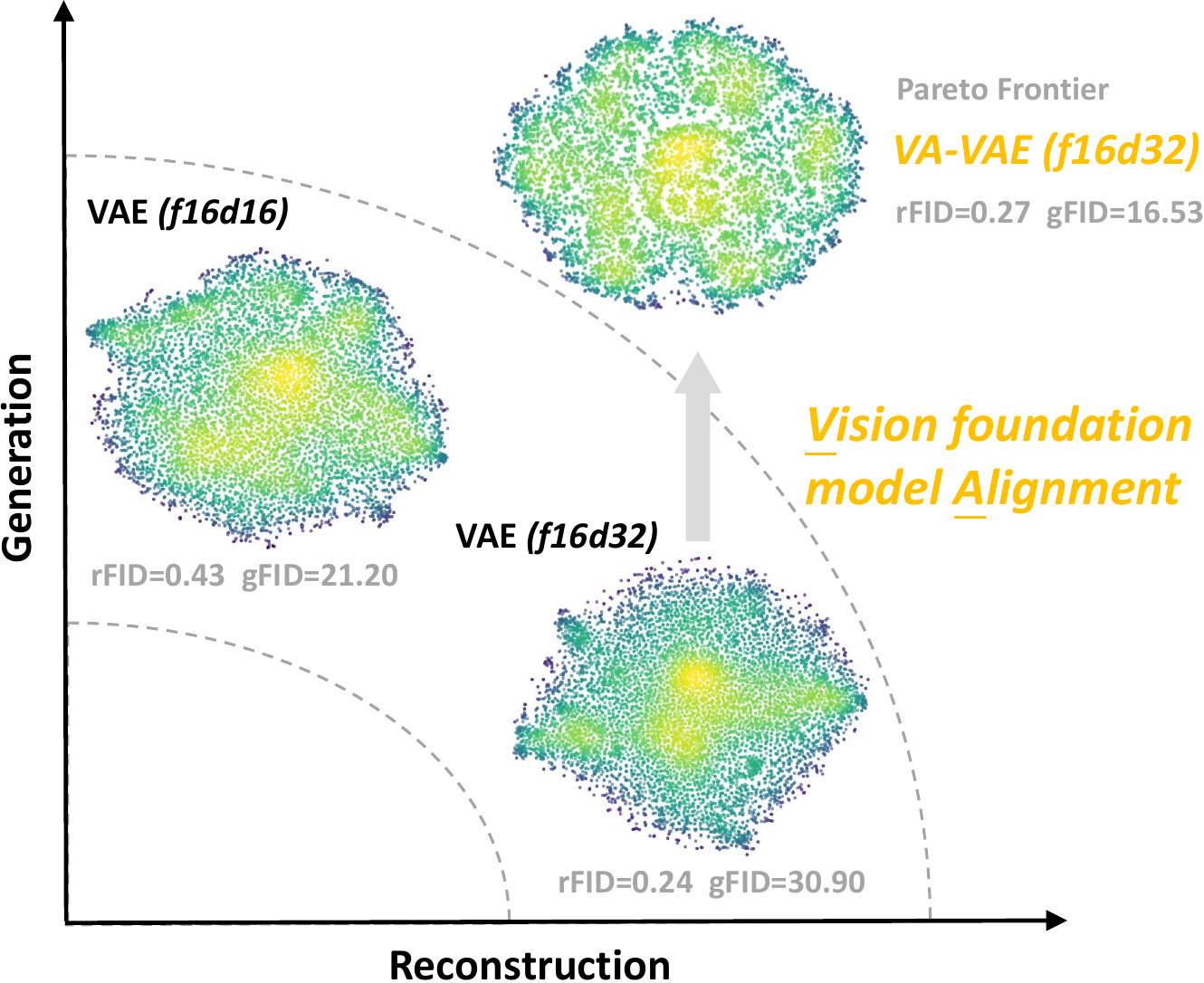}
    \caption{\textbf{Reconstruction-generation frontier of latent diffusion models.} VA-VAE improves the feature distribution of high-dimensional latent. Through alignment with vision foundation models, we expand the frontier between reconstruction and generation in latent diffusion models.}
    \label{fig:pareto_frontier}
\end{figure}

In this paper, we propose a straightforward yet effective approach to this optimization dilemma. We draw inspiration from Auto-Regressive (AR) generation, where increasing the codebook size of discrete-valued VAEs leads to low codebook utilization~\cite{magvitv2, lcvqgan}. Through visualizing the latent space distributions across different feature dimensions (see Fig.~\ref{fig:optimization_conflict}), we observe that higher-dimensional tokenizers learn latent representations in a less spread-out manner, evidenced by more concentrated high-intensity areas in the distribution visualization. This analysis suggests that the optimization dilemma stems from the inherent difficulty of learning unconstrained high-dimensional latent spaces from scratch. To address this issue, we develop a vision foundation model guided optimization strategy for continuous VAEs~\cite{vae} in latent diffusion models. Our key finding demonstrates that learning latent representations guided by vision foundation models significantly enhances the generation performance of high-dimensional tokenizers while preserving their original reconstruction capabilities (see Fig.~\ref{fig:pareto_frontier}).

Our main technical contribution is the Vision Foundation model alignment Loss (\textbf{VF Loss}), a plug-and-play module that aligns latent representations with pre-trained vision foundation models~\cite{dinov2, mae} during tokenizer training. While naively initializing VAE encoders with pre-trained vision foundation models has proven ineffective~\cite{pgv3}—likely because the latent representation quickly deviates from its initial state to optimize reconstruction—we find that a carefully designed joint reconstruction and alignment loss is crucial. Our alignment loss is specifically crafted to regularize high-dimensional latent spaces without overly constraining their capacity. First, we enforce both element-wise and pair-wise similarities to ensure comprehensive regularization of global and local structures in the feature space. Second, we introduce a margin in the similarity cost to provide controlled flexibility in the alignment, thereby preventing over-regularization. Additionally, we investigate the impact of different vision foundation models. 

To evaluate the generation performance, we couple the proposed Vision foundation model Aligned VAE (\textbf{VA-VAE}) with Diffusion Transformers (DiT)~\cite{dit} to create a latent diffusion model. To fully exploit the potential of VA-VAE, we build an enhanced DiT framework through advanced diffusion training strategies and transformer architectural improvements, which we name \textbf{LightningDiT}. Our contributions achieve the following significant milestones:

\begin{itemize}
    \item The proposed VF Loss effectively resolves the optimization dilemma in latent diffusion models, enabling over 2.5$\times$ faster DiT training with high-dimensional tokenizers;
    \item The integrated system reaches an FID of 2.11 within just 64 training epochs, an over 21$\times$ convergence speedup compared with the original DiT;
    \item The integrated system achieves a state-of-the-art (SOTA) FID score of 1.35 on ImageNet-256 image generation.
\end{itemize}

\begin{figure}
    \centering
    \includegraphics[width=0.8\linewidth]{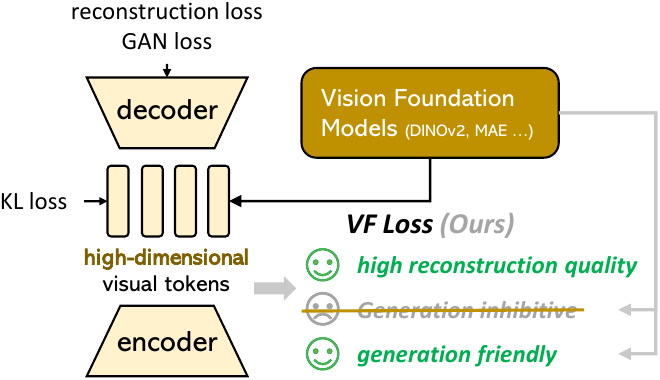}
    \caption{\textbf{The proposed Vision foundation model Aligned VAE (VA-VAE).} Vision foundation models are used to guide the training of high-dimensional visual tokenizers, effectively mitigating the optimization dilemma and improve generation performance.}
    \label{fig:simple_framework}
\end{figure}

\section{Related Work}

\subsection{Tokenizers for Visual Generation}

Visual tokenizers, represented by variational autoencoders (VAEs)~\cite{vae}, leverage an encoder-decoder architecture to create continuous visual representations, facilitating the compression and reconstruction of visual signals. While VAEs operate in continuous space, VQVAE~\cite{vqvae} introduces discrete representation learning through a learnable codebook for quantization. VQGAN~\cite{vqgan} further enhances this discrete approach by incorporating adversarial training, establishing itself as a cornerstone for autoregressive (AR) generation frameworks~\cite{maskgit}. However, these discrete approaches face a fundamental challenge: as revealed in~\cite{magvitv2}, larger codebooks improve reconstruction fidelity but lead to poor codebook utilization, adversely affecting generation performance. Recent works propose different solutions to this utilization problem: SeQ-GAN~\cite{seq-gan} explores the role of semantic information in generation. MAGVIT-v2~\cite{magvitv2} introduces Look-up Free Quantization (LFQ), while VQGAN-LC~\cite{lcvqgan} leverages CLIP~\cite{clip} vision features for codebook initialization, achieving near-complete codebook utilization.

Interestingly, continuous VAE-based latent diffusion systems~\cite{ldm} encounter a parallel optimization challenge: increasing the tokenizer's feature dimensionality enhances reconstruction quality but degrades generation performance, necessitating substantially higher training costs~\cite{sd3, dcae, pgv3}. Despite the significance of this trade-off in both discrete and continuous domains, current literature lacks comprehensive analysis and effective solutions for continuous VAE optimization. 
Our work addresses this fundamental limitation by introducing vision foundation model alignment into continuous VAE training. This principled approach resolves the optimization dilemma by structuring the latent space according to well-established visual representations, enabling efficient training of diffusion models in higher-dimensional latent spaces.
Importantly, our solution maintains the computational efficiency of diffusion model training as it requires no additional parameters, while significantly accelerating convergence by over 2.5 $\times$.

\subsection{Fast Convergence of Diffusion Transformers}

Diffusion Transformers~\cite{dit} are currently the most popular implementation of latent diffusion models~\cite{sd3, pixart-alpha, vdt, openai2024sora, lumina}. Due to their remarkable scalability, they have proven effective across various text-to-image and text-to-video tasks. However, they suffer from slow convergence speeds. Previous works have proposed various solutions: SiT~\cite{sit} enhances DiT's efficiency through Rectified Flow integration, while MDT~\cite{mdt} and MaskDiT~\cite{maskdit} achieve faster convergence by incorporating mask image modeling. REPA~\cite{repa} accelerates convergence by aligning DiT features with vision foundation models like DINOv2~\cite{dinov2} during training.

Different from these approaches, we identify that a major bottleneck in efficient latent diffusion training lies in the tokenizer itself. We propose a principled approach that optimizes the latent space learned by the visual tokenizer. Unlike methods that combine diffusion loss with auxiliary losses~\cite{mdt,maskdit,repa}, which incur additional computational costs during training, our approach achieves faster convergence without modifying the diffusion models. Additionally, we add several optimizations in terms of training strategies and architecture designs to the original DiT implementation that further accelerate training. 

\textit{\textsuperscript{*}Relationship to REPA~\cite{repa}}: Both our work and REPA~\cite{repa} utilize vision foundation models to aid in diffusion training, yet the motivations and approaches are entirely different. REPA aims to employ vision foundation models to constrain DiT, thereby \textit{enhancing the convergence speed} of generative models. In contrast, our work takes into account both the reconstruction and generative capabilities within the latent diffusion model, with the objective of leveraging foundation models to regulate the high-dimensional latent space of the tokenizer, thereby \textit{resolving the optimization conflict} between the tokenizer and the generative model.

\section{Align VAE with Vision Foundation Models}
\label{sec:vf_loss}

In this section, we introduce \thevae{}, a visual tokenizer trained through vision foundation model alignment. The key approach involves constraining the tokenizer's latent space by leveraging the feature space of the foundation model, which enhances its suitability for generative tasks. 

As illustrated in Figure~\ref{fig:simple_framework}, our architecture and training process mainly follows LDM~\cite{ldm}, employing a VQGAN~\cite{vqgan} model architecture with a continuous latent space, constrained by KL loss. Our key contribution lies in the design of \textbf{V}ision \textbf{F}oundation model alignment loss, \textbf{VF loss}, which substantially optimizes the latent space without altering the model architecture or training pipeline, effectively resolving the optimization dilemma discussed in Section~\ref{sec:intro}.

The VF loss consists of two components: marginal cosine similarity loss and marginal distance matrix similarity loss. These components are carefully crafted as a straightforward, plug-and-play module that is decoupled from the VAE architecture. We will explain each part in detail below.

\subsection{Marginal Cosine Similarity Loss}

During training, a given image \( I \) is processed by both the encoder of visual tokenizer and a frozen vision foundation model, resulting in image latents \( Z \) and foundational visual representations \( F \). As shown in Eq.~\ref{eq:feature_projection}, we project \( Z \) to match the dimensionality of \( F \) using a linear transformation, where \( W \in \mathbb{R}^{d_f \times d_z} \) respectively, producing \( Z' \in \mathbb{R}^{d_f} \). 
\begin{equation}
\label{eq:feature_projection}
    Z' = W Z
\end{equation}

As defined in Eq.~\ref{eq:margin_cos}, the loss function \(\mathcal{L}_{\text{mcos}}\) minimizes the similarity gap between corresponding features \( z'_{ij} \) and \( f_{ij} \) from feature matrices \( Z' \) and \( F \) at each spatial location \((i, j)\). For each pair, we compute the cosine similarity \( \frac{z'_{ij} \cdot f_{ij}}{\|z'_{ij}\| \|f_{ij}\|} \) and subtract a margin \( m_1 \). The \(\text{ReLU}\) function ensures that only pairs with similarities below \( m_1 \) contribute to the loss, focusing alignment on less similar pairs. The final loss is averaged over all positions in the \( h \times w \) feature grid.

\begin{equation}
\label{eq:margin_cos}
\mathcal{L}_{\text{mcos}} = \frac{1}{h \times w} \sum_{i=1}^{h} \sum_{j=1}^{w} \text{ReLU} \left( 1 - m_1 - \frac{z'_{ij} \cdot f_{ij}}{\|z'_{ij}\| \|f_{ij}\|} \right)
\end{equation}

\subsection{Marginal Distance Matrix Similarity Loss}

Complementary to \(\mathcal{L}_{\text{mcos}}\), which enforces point-to-point absolute alignment, we also aim for the relative distribution distance matrices within the features to be as similar as possible. To achieve this, we propose the marginal distance matrix similarity loss.

In Eq.~\ref{eq:dist_matrix_sim_margin}, the distance matrix similarity Loss aligns the internal distributions of feature matrices \( z \) and \( f \). Here, \( N = h \times w \) represents the total number of elements in each flattened feature map. For each pair \( (i, j) \), we compute the absolute value of the cosine similarity difference between the corresponding vectors in feature matrices \( z \) and \( f \), thus promoting closer alignment of their relative structures. Similarly, we subtract a margin \( m_2 \) to relax the constraint. The \(\text{ReLU}\) function ensures that only pairs with differences exceeding \( m_2 \) contribute to the loss.

\begin{equation}
\label{eq:dist_matrix_sim_margin}
\mathcal{L}_{\text{mdms}} = \frac{1}{N^2} \sum_{i,j} \text{ReLU} \left(\left| \frac{z_i \cdot z_j}{\|z_i\| \|z_j\|} - \frac{f_i \cdot f_j}{\|f_i\| \|f_j\|} \right| - m_2 \right)
\end{equation}

\subsection{Adaptive Weighting}

In Figure~\ref{fig:simple_framework}, the original reconstruction loss and KL loss are both sum losses, which places the VF loss on a completely different scale, making it challenging to adjust the weight for stable training. Inspired by GAN Loss~\cite{vqgan}, we employ an adaptive weighting mechanism. Before backpropagation, we calculate the gradients of \( L_{\text{vf}} \) and \( L_{\text{rec}} \) on the last convolutional layer of the encoder, as shown in Eq~\ref{eq:adaptive_weighting}. The adaptive weighting is set as the ratio of these two gradients, ensuring that \( L_{\text{vf}} \) and \( L_{\text{rec}} \) have similar impacts on model optimization. This alignment significantly reduces the adjustment range of the VF Loss.

\begin{equation}
\label{eq:adaptive_weighting}
w_{\text{adaptive}} = \frac{\|\nabla L_{\text{rec}}\|}{\|\nabla L_{\text{vf}}\|}
\end{equation}

Then we get VF Loss with adaptive weighting like Eq~\ref{eq:vf_loss}. The purpose of adaptive weighting is to quickly align our loss scales across different VAE training pipelines. On this basis, we can still use manually tuned hyperparameters to further improve performance.

We will evaluate the significant role of VF loss in achieving the latent diffusion Pareto frontier for both reconstruction and generation in our forthcoming experiments.

\begin{equation}
\label{eq:vf_loss}
\mathcal{L}_{\text{vf}} = w_{\text{hyper}} * w_{\text{adaptive}}(\mathcal{L}_{\text{mcos}} + \mathcal{L}_{\text{mdms}})
\end{equation}

\section{Improved Diffusion Transformer}
\label{sec:veryfastdit}

\begin{table}[t]
\centering
\resizebox{\linewidth}{!}{
\begin{tabular}{llcr}
\toprule
\textbf{Training Trick} & \textbf{Training Sample} & \textbf{Epoch} & \textbf{FID-50k}$\downarrow$ \\ 
\midrule
DiT-XL/2~\cite{dit} & 400k $\times$ 256 & 80 & 19.50 \\
\midrule
\multicolumn{4}{c}{\textbf{\textit{Training Strategies}}} \\
\midrule
+ Rectified Flow~\cite{rf} & 400k $\times$ 256  & \multirow{5}{*}{80} & 17.20 \\
+ \textit{batchsize}$\times4$ \& \textit{lr}$\times2$ & \multirow{4}{*}{100k $\times$ 1024} &  & 16.59 \\
+ AdamW $\beta_2=0.95$~\cite{beta2} &  &  & 16.61 \\
+ Logit Normal Sampling~\cite{sd3} &  &  & 13.99 \\
+ Velocity Direction Loss~\cite{fasterdit}  & &  & 12.52 \\
\midrule
\multicolumn{4}{c}{\textbf{\textit{Architecture Improvements}}} \\
\midrule
+ SwiGLU FFN~\cite{swiglu} & \multirow{4}{*}{100k $\times$ 1024} & \multirow{4}{*}{80} & 10.10 \\
+ RMS Norm~\cite{rmsnorm} &  &  & 9.25 \\
+ Rotary Pos Embed~\cite{rope} & &  & 7.13 \\
+ patch size=1 \& VA-VAE (Sec.~\ref{sec:vf_loss}) & &  & 4.29 \\
\bottomrule
\end{tabular}}
\caption{\textbf{Performance of \thedit{}}. With SD-VAE~\cite{ldm}, \thedit~achieves FID-50k=7.13 on ImageNet class-conditional generation, using 94\% fewer training samples compared to the original DiT~\cite{dit}. 
We show that the original DiT can also achieve exceptional performance by leveraging advanced design techniques.}
\label{tab:veryfastdit}
\end{table}

\begin{table*}[t]
\centering
\renewcommand{\arraystretch}{1.1}
\resizebox{\textwidth}{!}{
\begin{tabular}{lccccc|ccc}
\toprule
\multirow{2}{*}{\textbf{Tokenizer}} & \multirow{2}{*}{\textbf{Spec.}} & \multicolumn{4}{c|}{\textbf{Reconstruction Performance}} & \multicolumn{3}{c}{\textbf{Generation Performance (FID-10K)$\downarrow$}} \\
& & \textbf{rFID}$\downarrow$ & \textbf{PSNR$\uparrow$} & \textbf{LPIPS$\downarrow$} & \textbf{SSIM$\uparrow$} & \textbf{LightningDiT-B} & \textbf{LightningDiT-L} & \textbf{LightningDiT-XL} \\
\midrule
LDM~\cite{ldm} & \multirow{3}{*}{f16d16} & \cellcolor{blue!20}0.49 & 26.10 & 0.132 & 0.72 & \multicolumn{1}{c}{\cellcolor{blue!20}16.24} & 9.49 & 8.28 \\
LDM+VF loss (\textit{MAE})~\cite{mae} & & 0.51 & 26.01 & 0.137 & 0.71 & 16.86 \textcolor{gray}{\textit{(+0.62)}} & 10.93 \textcolor{gray}{\textit{(+1.44)}} & 9.19 \textcolor{gray}{\textit{(+0.91)}} \\
LDM+VF loss (\textit{DINOv2})~\cite{dinov2} & & 0.55 & 25.29 & 0.147 & 0.69 & 15.79 \textcolor{orange}{\textit{(-0.45)}} & 10.02 \textcolor{gray}{\textit{(+0.53)}} & 8.71 \textcolor{gray}{\textit{(+0.43)}} \\
\midrule
LDM~\cite{ldm} & \multirow{3}{*}{f16d32} & \cellcolor{blue!20}0.26 & 28.59 & 0.089 & 0.80 & \multicolumn{1}{c}{\cellcolor{blue!20}22.62} & 12.86 & 10.92 \\
LDM+VF loss (\textit{MAE})~\cite{mae} & & 0.28 & 28.33 & 0.091 & 0.80 & 19.89 \textcolor{orange}{\textit{(-2.73)}} & 11.51 \textcolor{orange}{\textit{(-1.35)}} & 9.92\phantom{0} \textcolor{orange}{\textit{(-1.00)}} \\
LDM+VF loss (\textit{DINOv2})~\cite{dinov2} & & 0.28 & 27.96 & 0.096 & 0.79 & 15.82 \textcolor{orange}{\textit{(-6.80)}} & 9.82\phantom{0} \textcolor{orange}{\textit{(-3.04)}} & 8.22\phantom{0} \textcolor{orange}{\textit{(-2.70)}} \\
\midrule
LDM~\cite{ldm} & \multirow{3}{*}{f16d64} & \cellcolor{blue!20}0.17 & 31.03 & 0.055 & 0.88 & \multicolumn{1}{c}{\cellcolor{blue!20}36.83} & 20.73 & 17.24 \\
LDM+VF loss (\textit{MAE})~\cite{mae} & & 0.15 & 31.03 & 0.054 & 0.87 & 23.58 \textcolor{orange}{\textit{(-13.25)}} & 14.40 \textcolor{orange}{\textit{(-6.33)}} & 11.69 \textcolor{orange}{\textit{(-5.55)}} \\
LDM+VF loss (\textit{DINOv2})~\cite{dinov2} & & 0.14 & 30.71 & 0.055 & 0.87 & 24.00 \textcolor{orange}{\textit{(-12.83)}} & 14.95 \textcolor{orange}{\textit{(-5.78)}} & 11.98 \textcolor{orange}{\textit{(-5.26)}} \\
\bottomrule
\end{tabular}}
\caption{\textbf{VF loss Improves Generation Performance.} The \textit{f16d16} tokenizer specification is widely used~\cite{ldm, mar}. As dimensionality increases, we observe that (1) higher dimensions improve reconstruction but reduce generation quality, highlighting an optimization dilemma within the latent diffusion framework; (2) VF Loss significantly enhances generative performance in high-dimensional tokenizers with minimal impact on reconstruction.}
\label{tab:vf_improve_convergence}
\end{table*}

In this section, we introduce our \thedit{}.
Diffusion Transformers (DiT)~\cite{dit} has gained considerable success as a foundational model for text-to-image~\cite{pixart-alpha, sd3} and text-to-video generation tasks~\cite{openai2024sora, moviegen}. However, its convergence speed on ImageNet is significantly slow, resulting in high experimental iteration costs. 
Previous influential work like DINOv2~\cite{dinov2}, ConvNeXt~\cite{convnext}, and EVA~\cite{eva} demonstrates how incorporating advanced design strategies can revitalize classic methods~\cite{ibot, resnet}. In our work, we aim to extend the potential of the DiT architecture and explore the boundaries of how far the DiT can go. While we do not claim any individual optimization detail as our original contribution, we believe an open-source, fast-converging training pipeline for DiT will greatly support the community’s ongoing research on diffusion transformers.

We utilize the SD-VAE~\cite{ldm} with the \textit{f8d4} specification as the visual tokenizer and employ DiT-XL/2 as our experimental model. We show the optimization routine in Table~\ref{tab:veryfastdit}. 
Each model has been trained for 80 epochs and sampled with a dopri5 integrator, which has less NFE than the original DiT for fast inference. To ensure a fair comparison, no sample optimization methods such as cfg interval~\cite{cfg_interval} and timestep shift~\cite{flux} are used.
We adopt three categories of optimization strategies. At the computational level, we implement \textit{torch.compile}~\cite{torch_compile} and \textit{bfloat16} training for acceleration. Additionally, we increase the batch size and reduce the $\beta_2$ of AdamW to 0.95, drawing from previous work AuraFlow~\cite{auraflow}. For diffusion optimization, we incorporate Rectified Flow~\cite{rf, sit}, logit normal distribution (lognorm) sampling~\cite{sd3}, and velocity direction loss~\cite{fasterdit}. At the model architecture level, we apply common Transformer optimizations, including RMSNorm~\cite{rmsnorm}, SwiGLU~\cite{swiglu}, and RoPE~\cite{rope}. During training, we observe that some acceleration strategies are not orthogonal. For example, gradient clipping is effective when used alone but tends to reduce performance when combined after lognorm sampling and velocity direction loss.

Our optimized model, \thedit, achieves an FID of 7.13 (\textit{cfg=1}) with SD-VAE at around 80 epochs on ImageNet class-conditional generation, which is only 6\% of the training volume required by the original DiT and SiT models over 1400 epochs. Previous great work MDT~\cite{mdt} or REPA~\cite{repa} achieved similar convergence performance with the help of Mask Image Modeling (MIM) and representation alignment. Our results show that, even without any complex training pipeline, naive DiT could still achieve very competitive performance. This optimized architecture has been of great help in our following rapid experiment validation.

\section{Experiments}
\label{sec:experiments}

\begin{figure*}
    \centering
    \begin{subfigure}[b]{0.31\linewidth}
        \includegraphics[width=\linewidth]{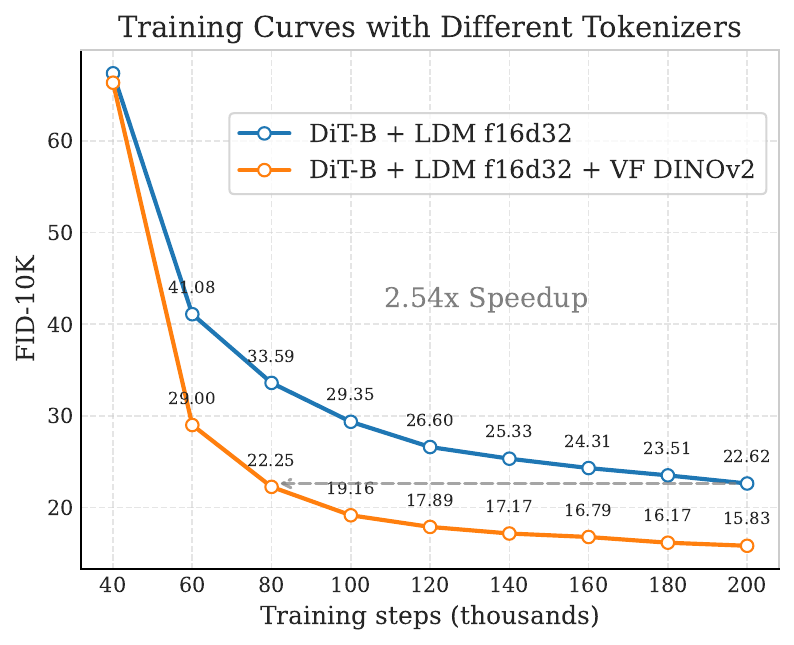}
        \caption{}
        \label{fig:curve_a}
    \end{subfigure}
    \hfill
    \begin{subfigure}[b]{0.31\linewidth}
        \includegraphics[width=\linewidth]{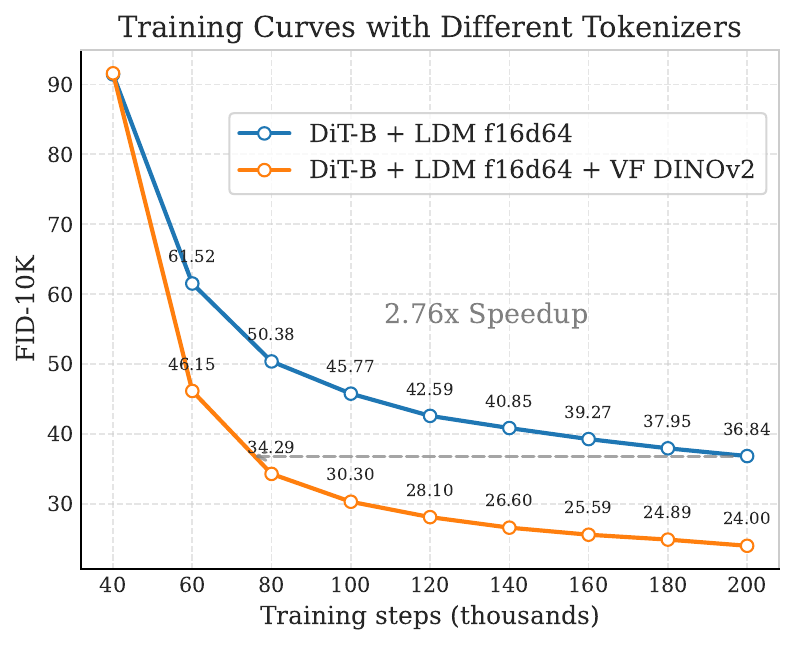}
        \caption{}
        \label{fig:curve_b}
    \end{subfigure}
    \hfill
    \begin{subfigure}[b]{0.333\linewidth}
        \includegraphics[width=\linewidth]{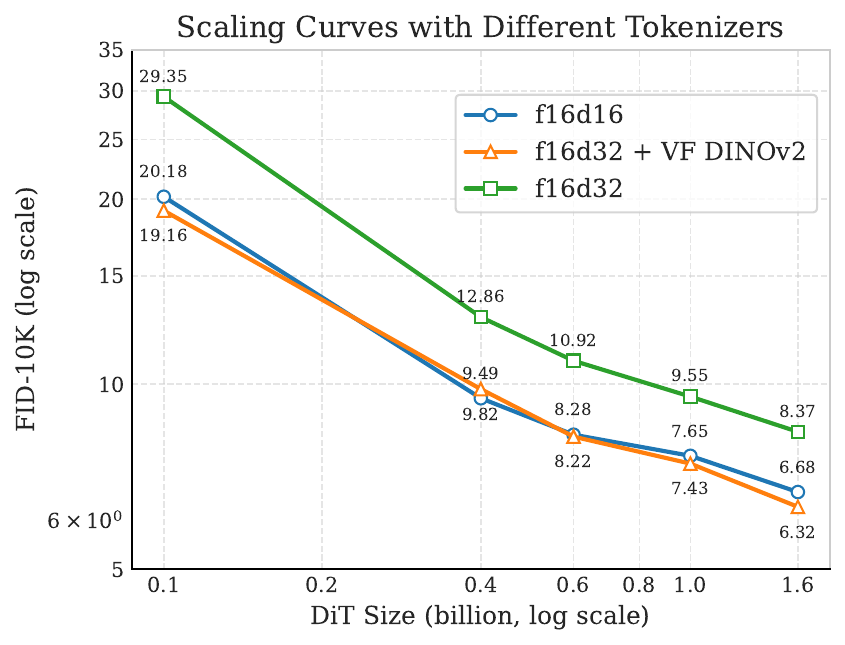}
        \caption{}
        \label{fig:curve_c}
    \end{subfigure}
    \caption{\textbf{(a)\&(b) VF Loss Improves Convergence.} We train \thedit{}-B for 160 epochs on ImageNet at 256 resolution using different tokenizers. The VF loss significantly accelerates convergence, with a maximum speedup of up to 2.7 times. \textbf{(c) VF Loss Improves Scalability.} VF loss reduces the need for large parameters in generative models of high-dimensional tokenizer, enabling better scalability.}
    \label{fig:convergence_and_scaling}
\end{figure*}


In this section, our main objective is to achieve the reconstruction and generation frontier (see Figure~\ref{fig:pareto_frontier}) of reconstruction and generation within the latent diffusion system by leveraging VF loss proposed in Section~\ref{sec:vf_loss}. With the support of \thedit{} introduced in Section~\ref{sec:veryfastdit}, we demonstrate how VF loss effectively resolves the optimization dilemma, from the perspective of convergence, scalability, and overall system performance.

\subsection{Implementation Details}

We introduce our latent diffusion system in detail.
For the visual tokenizer, we employ an architecture and training strategy mainly following to LDM~\cite{ldm}. Specifically, we utilize the VQGAN~\cite{vqgan} network structure, omitting quantization and applying KL Loss to regulate the continuous latent space. To enable multi-node training, we scale the learning rate and global batch size to $1e-4$ and $256$, respectively, following a setup from MAR~\cite{mar}. We train three different $f16$ tokenizers: one without VF loss, one using VF loss (\textit{MAE}), and another using VF loss (\textit{DINOv2}). Here $f$ denotes the downsampling rate and $d$ denotes the latent dimension. Empirically, we set $m_1=0.5$, $m_2=0.25$, and $w_{hyper}=0.1$. We argue different vision foundation models might converge to different margin settings. For the generative model, we employ \thedit{}, which is further refined with the design techniques outlined in Section~\ref{sec:veryfastdit}. We pre-extract all latent features from the tokenizer and train various versions of \thedit{} on ImageNet at a resolution of 256 for either 80 or 160 epochs. We set the patch size of DiT to 1, ensuring that the downsampling rate of the entire system is 16.  This approach is consistent with the strategy proposed in \cite{dcae}, i.e. all compression steps are handled by the VAE. Unless otherwise noted, our model's other architectural parameters are consistent with those of DiT~\cite{dit}.

\begin{table*}[t]
\centering
\resizebox{\textwidth}{!}{
\begin{tabular}{l|cc|ccccccc|ccccc}
\toprule
\multirow{2}{*}{\textbf{Method}} & \multicolumn{2}{c|}{\textbf{Reconstruction}} & \multirow{2}{*}{\makecell{\textbf{Training} \\ \textbf{Epoches}}} & \multirow{2}{*}{\textbf{\#params}} & \multicolumn{5}{c|}{\textbf{Generation w/o CFG}} & \multicolumn{5}{c}{\textbf{Generation w/ CFG}} \\
\cmidrule(lr){6-10} \cmidrule(lr){11-15}
 & \textbf{Tokenizer} & \textbf{rFID} & & & \textbf{gFID} & \textbf{sFID} & \textbf{IS} & \textbf{Pre.} & \textbf{Rec.} & \textbf{gFID} & \textbf{sFID} & \textbf{IS} & \textbf{Pre.} & \textbf{Rec.} \\
\midrule
& \multicolumn{11}{c}{\textit{\textbf{AutoRegressive (AR)}}} \\
\midrule
MaskGIT~\cite{maskgit} & MaskGiT & 2.28 & 555 & 227M & 6.18 & - & 182.1 & 0.80 & 0.51 & -  & - & -  & - & - \\
LlamaGen~\cite{llamagen}  & VQGAN\textsuperscript{\textcolor{purple}{†}} & 0.59 & 300 & 3.1B & 9.38 & 8.24 & 112.9 & 0.69& 0.67 & 2.18 & 5.97 & 263.3 & 0.81 & 0.58 \\
VAR~\cite{var} & - & - & 350 & 2.0B & - & - & - & - & - & 1.80 & - & 365.4 & 0.83 & 0.57 \\
MagViT-v2~\cite{magvitv2} & - & - & 1080 & 307M & 3.65 & - & 200.5 & - & - & 1.78 & - & 319.4 & - & - \\
MAR~\cite{mar} & LDM\textsuperscript{\textcolor{purple}{†}} & 0.53 & 800 & 945M & 2.35 & - & 227.8 & 0.79 & 0.62 & 1.55 & - & 303.7 & 0.81 & 0.62 \\
\midrule
& \multicolumn{11}{c}{\textit{\textbf{Latent Diffusion Models}}} \\
\midrule
MaskDiT~\cite{maskdit} & \multirow{6}{*}{SD-VAE~\cite{ldm}} & \multirow{6}{*}{0.61} & 1600 & 675M & 5.69 & 10.34 & 177.9 & 0.74 & 0.60 & 2.28 & 5.67 & 276.6 & 0.80 & 0.61 \\ 
DiT~\cite{dit} & & & 1400 & 675M & 9.62 & 6.85 & 121.5 & 0.67 & 0.67 & 2.27 & 4.60 & 278.2 & \textbf{0.83} & 0.57 \\
SiT~\cite{sit} & & & 1400 & 675M& 8.61 & 6.32 & 131.7 & 0.68 & 0.67 & 2.06 & 4.50 & 270.3 & 0.82 & 0.59 \\
FasterDiT~\cite{fasterdit} & & & 400 & 675M & 7.91 & 5.45 & 131.3 & 0.67 & \textbf{0.69} & 2.03 & 4.63 & 264.0 & 0.81 & 0.60 \\
MDT~\cite{mdt} & & & 1300 & 675M & 6.23 & 5.23 & 143.0 & 0.71 & 0.65 & 1.79 & 4.57 & 283.0 & 0.81 & 0.61 \\
MDTv2~\cite{mdtv2} & & & 1080 & 675M & - & - & - & - & - &1.58 & 4.52 & \textbf{314.7} & 0.79 & 0.65 \\ 
REPA~\cite{repa} & & & 800 & 675M & 5.90 & - & - & - & - & 1.42 & 4.70 & 305.7 & 0.80 & 0.65 \\
\midrule
\multirow{2}{*}{\textbf{\thedit}} & \multirow{2}{*}{\textbf{\thevae}} & \multirow{2}{*}{\textbf{0.28}} & \cellcolor{blue!20}\textcolor{gray}{\textbf{64}} & \cellcolor{blue!20}\textcolor{gray}{675M} & \cellcolor{blue!20}\textcolor{gray}{5.14} & \cellcolor{blue!20}\textcolor{gray}{\textbf{4.22}} & \cellcolor{blue!20}\textcolor{gray}{130.2} & \cellcolor{blue!20}\textcolor{gray}{0.76}  & \cellcolor{blue!20}\textcolor{gray}{0.62} & \cellcolor{blue!20}\textcolor{gray}{2.11} & \cellcolor{blue!20}\textcolor{gray}{4.16} & \cellcolor{blue!20}\textcolor{gray}{252.3} & \cellcolor{blue!20}\textcolor{gray}{0.81} & \cellcolor{blue!20}\textcolor{gray}{0.58} \\
 &  & & \cellcolor{blue!20}800 & \cellcolor{blue!20}675M & \cellcolor{blue!20}\textbf{2.17} & \cellcolor{blue!20}4.36 & \cellcolor{blue!20}\textbf{205.6} & \cellcolor{blue!20}\textbf{0.77} & \cellcolor{blue!20}0.65 & \cellcolor{blue!20}\textbf{1.35} & \cellcolor{blue!20}\textbf{4.15}  & \cellcolor{blue!20}295.3 & \cellcolor{blue!20}0.79 & \cellcolor{blue!20}\textbf{0.65} \\
\bottomrule
\end{tabular}}
\caption{\textbf{System-Level Performance on ImageNet 256$\times$256.} Our latent diffusion system achieves \textbf{\textit{state-of-the-art}} performance with rFID=0.28 and FID=1.35. Besides, our LightningDiT together with VA-VAE surpasses DiT~\cite{dit} and SiT~\cite{sit} in FID within only 64 training epochs, demonstrating a \textbf{\textit{21.8 $\times$ faster convergence.}}}
\label{tab:method_comparison}
\end{table*}

\begin{figure*}[t]
    \centering
    \includegraphics[width=\linewidth]{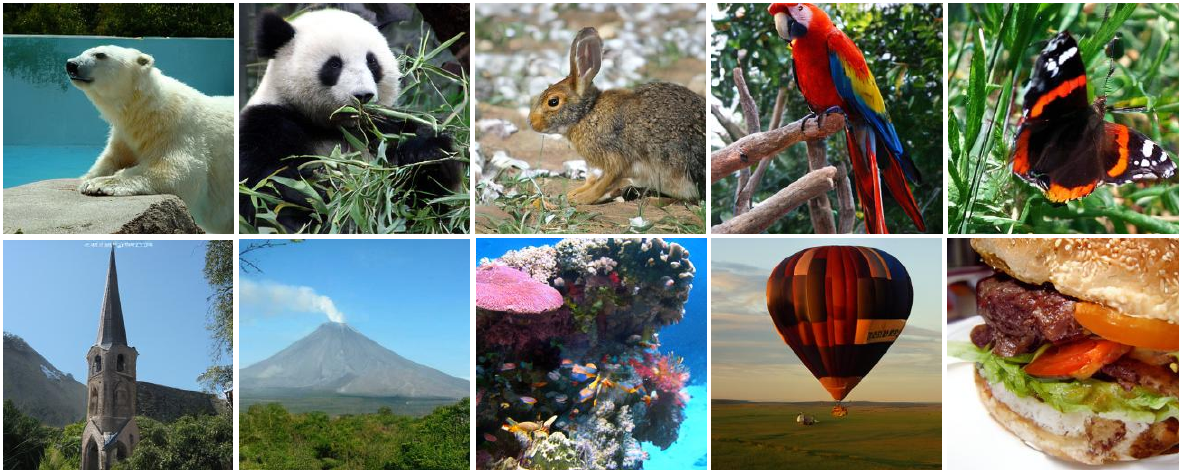}
    \caption{\textbf{Visualization Results.} We visualize our latent diffusion system with proposed \thevae{} together with \thedit{}-XL trained on ImageNet $256\times256$ resolution. }
    \label{fig:enter-label}
\end{figure*}

\subsection{Foundation Models Improve Convergence}

Table~\ref{tab:vf_improve_convergence} presents an evaluation of the reconstruction and generation of eight different tokenizers, with all generative models trained for 160 epochs (LightningDiT-B) or 80 epochs (LightningDiT-L \& LightningDiT-XL) on ImageNet. Here we come with the following findings: 

The results highlight the optimization dilemma in latent diffusion systems, as discussed in Section~\ref{sec:intro}. The results highlighted in blue in the table illustrate the reconstruction performance (rFID) and the corresponding generation performance (FID). It can be observed that as the tokenizer dimension increases, its rFID decreases, while the corresponding generation FID increases.

The VF Loss can effectively enhance the generative performance of high-dimensional tokenizers. In the f16d32 and f16d64 sections of the table, both VF loss (\textit{DINOv2}) and VF loss (\textit{MAE}) significantly improve the generative performance of DiT models across different scales. This makes it possible to achieve systems with higher reconstruction performance and higher generative performance (i.e., the reconstruction-generation frontier mentioned in the introduction). It is worth noting, however, that the VF loss is unnecessary for lower-dimensional tokenizers, such as normally used f16d16~\cite{ldm, mar, llamagen}. This stays consistent with the latent distribution observation in Figure~\ref{fig:optimization_conflict}. 
We suggest this is because lower-dimensional spaces can learn more reasonable distributions without the need for additional supervisory signals.

Additionally, we present a convergence plot of FID over training time in Figure~\ref{fig:convergence_and_scaling} (a) \& (b). On f16d32 and f16d64, the use of VF loss accelerates convergence by factors of 2.54 and 2.76, respectively. These also demonstrate that the VF loss significantly enhances the generative performance and convergence speed of high-dimensional tokenizers.

\subsection{Foundation Models Improve Scalability}

As discussed in Section~\ref{sec:intro}, increasing the model parameter count serves as a way to improve the generative performance of high-dimensional tokenizers~\cite{sd3}. We use \thedit{} models ranging from 0.1B to 1.6B in size to evaluate the generative performance of 3 different tokenizers. 

To facilitate the observation of the power law in scaling, we employ a log scale for the axes. We notice a slight convergence trend between the blue and green lines as the number of parameters increases, yet a significant gap remains. This implies that the negative effects on generation brought by high-dimensional f16d32 tokenizers are not fully mitigated even at 1.6B, a parameter size already considered substantial on ImageNet. We find that the VF loss effectively bridges this gap. Below 0.6B, the performance of the orange and blue lines is similar. However, as the model scales beyond 1B, f16d32 VF DINOv2 gradually distances itself from f16d16, demonstrating stronger scalability.

\subsection{Convergence 21.8$\times$ Faster than DiT}
\label{sec:c2i_exp}

We find that the VF loss (\textit{DINOv2}) brings the most significant improvement in generative performance. Therefore, we extend the training time for the tokenizer and adopt a progressive training strategy to train the LDM VF loss (\textit{DINOv2}) for 125 epochs, resulting in a VA-VAE with stronger generative capabilities through prolonged training.
We train LightningDiT-XL for 800 epochs following the parameters in Table~\ref{tab:veryfastdit}. Specifically, at 480 epochs, we disable the lognorm parameter to enable the near-converged network to learn more effectively across all noise intervals. During sampling, we use a 250-step Euler integrator, ensuring the same NFE as previous works such as REPA~\cite{repa} and DiT~\cite{dit}. To enhance sampling performance, we adopt cfg interval~\cite{cfg_interval} and timestep shift similar to FLUX~\cite{flux}. We benchmark our method against prior AR generation and latent diffusion approaches in Table~\ref{tab:method_comparison}.

We report four distinct sets of results, detailing the performance with and without cfg for both extended training (800 epochs) and rapid training (64 epochs). At 800 epochs, our model achieves state-of-the-art performance with an FID of 1.35. Furthermore, our model demonstrates exceptional performance in cfg-free generation, achieving an FID of 2.17, which surpasses the results of many methods that utilize cfg. Our approach also exhibits rapid convergence capabilities; at 64 epochs, it achieves an FID of 2.11, representing a speedup of over 21 times compared to the original DiT. This further underscores the superiority of our method.

\section{Ablations and Discussions}

In this section, we perform ablation experiments on the design of VF loss to assess the impact of various foundation models and loss formulations. We then provide a deeper analysis of the underlying mechanism of VF loss, offering additional insights that might be helpful.

\subsection{Generative Friendly VA-VAE}

We demonstrate that the VA-VAE with a patch size of 1 exhibits superior generative performance compared to the SD-VAE with a patch size of 2. As shown in Table~\ref{tab:veryfastdit}, replacing the SD-VAE~\cite{ldm} with the VA-VAE results in a reduction of the FID-50k from 7.13 to 4.29. This improvement can be attributed to two main reasons. Firstly, we observe that the DiT trained with a tokenizer using f16 and a patch size of 1 converges more readily than the DiT using f8 and a patch size of 2. Secondly, the vision foundation model is capable of enhancing its generative performance while maintaining its reconstruction fidelity.

\subsection{Alations on Vision Foundation Models}

We train our \thevae{} using three types of foundation models: self-supervised models~\cite{mae, dinov2} with masked image modeling, the image-text contrastive learning model CLIP~\cite{clip}, and the Segment Anything model~\cite{sam}. As in Section~\ref{sec:experiments}, we set $w_{hyper}$ to 0.1, with $m_1=0.5$ and $m_2=0.25$. To accelerate convergence, we adjust the learning rate and global batch size to 1e-4 and 256, respectively. In contrast to previous settings, each tokenizer is trained on ImageNet $256\times256$ for 50 epochs. For each tokenizer, we train \thedit-B in the corresponding latent space for 160 epochs.
Table~\ref{tab:vision_models} summarizes our findings, showing that all these vision foundation models enhance the generative performance of diffusion models. Among them, the self-supervised pre-trained model DINOv2 achieves superior generative results. 

\subsection{Ablations on Loss Formulations}

We conduct ablation experiments on the loss functions proposed in Section~\ref{sec:vf_loss}. In these experiments, we use DINOv2 as the vision foundation model to train the \textit{f16d32} tokenizers for 50 epochs, comparing the reconstruction and generative results with different settings. We individually remove the margin cosine similarity loss (mcos), margin distance matrix similarity loss (mdms), and the margin from the loss function. Due to the presence of adaptive weighting, when we use a single loss individually, we halve the hyper weight to ensure a fair comparison. For all three scenarios, we observe a certain degree of performance degradation, which validates the effectiveness of these components.

\begin{table}[t]
\centering
\renewcommand{\arraystretch}{1.1}
\resizebox{\linewidth}{!}{ 
\begin{tabular}{l|cccc|c}
\toprule
\textbf{Model Type} & \textbf{rFID}$\downarrow$ & \textbf{PSNR}$\uparrow$ & \textbf{LPIPS}$\downarrow$ & \textbf{SSIM}$\uparrow$ & \textbf{gFID}$\downarrow$ \\ 
\midrule
naive & 0.26 & 28.59 & 0.089 & 0.80 & 22.62 \\
\midrule
DINOv2~\cite{dinov2} & 0.28 & 27.96 & 0.096 & 0.79 & \textbf{15.82} \\
MAE~\cite{mae} & 0.28 & 28.33 & 0.091 & 0.80 & 19.89 \\
SAM~\cite{sam} & 0.26 & 28.31 & 0.091 & 0.80 & 19.80 \\
CLIP~\cite{clip} & 0.33 & 28.39 & 0.091 & 0.80 & 18.93 \\
\bottomrule
\end{tabular}
}
\caption{\textbf{Ablation on Foundation Models.} We evaluate the impact of different VF losses on generative performance. Our results show that DINOv2 achieves the highest generative performance.}
\label{tab:vision_models}
\end{table}

\begin{table}[t]
\centering
\renewcommand{\arraystretch}{1.1}
\resizebox{\linewidth}{!}{
\begin{tabular}{l|cccc|c}
\toprule
\textbf{Loss Type} & \textbf{rFID}$\downarrow$ & \textbf{PSNR}$\uparrow$ & \textbf{LPIPS}$\downarrow$ & \textbf{SSIM}$\uparrow$ & \textbf{gFID}$\downarrow$ \\
\midrule
\textit{\textbf{NaN}} & 0.26 & 28.59 & 0.089 & 0.80 & 22.62 \\
\midrule
\textit{\textbf{full}} & 0.28 & 27.96 & 0.096 & 0.79 & \textbf{15.82} \\
\textit{-mcos loss} & 0.27 & 28.52 & 0.090 & 0.80 & 21.87 \\
\textit{-mdistmat loss} & 0.27 & 28.24 & 0.090 & 0.80 & 17.74 \\
\textit{-margin} & 0.27 & 28.07 & 0.093 & 0.79 & 17.77 \\
\bottomrule
\end{tabular}}
\caption{\textbf{Ablation Study of VF Loss Formulations}: 
Comparison of different configurations on generative performance metrics using LightningDiT-B.}
\label{tab:vf_loss_ablation}
\end{table}

\subsection{Discuss on VF loss with Latent Distribution}

In discrete visual tokenizers, there is also a conflict between reconstruction and generation~\cite{magvitv2, lcvqgan}. A clear indicator of this conflict is codebook utilization. When the codebook is scaled up, reconstruction performance improves, but codebook utilization significantly decreases, resulting in uneven distribution in the discrete space. 

\begin{figure}[t]
    \centering
    \includegraphics[width=\linewidth]{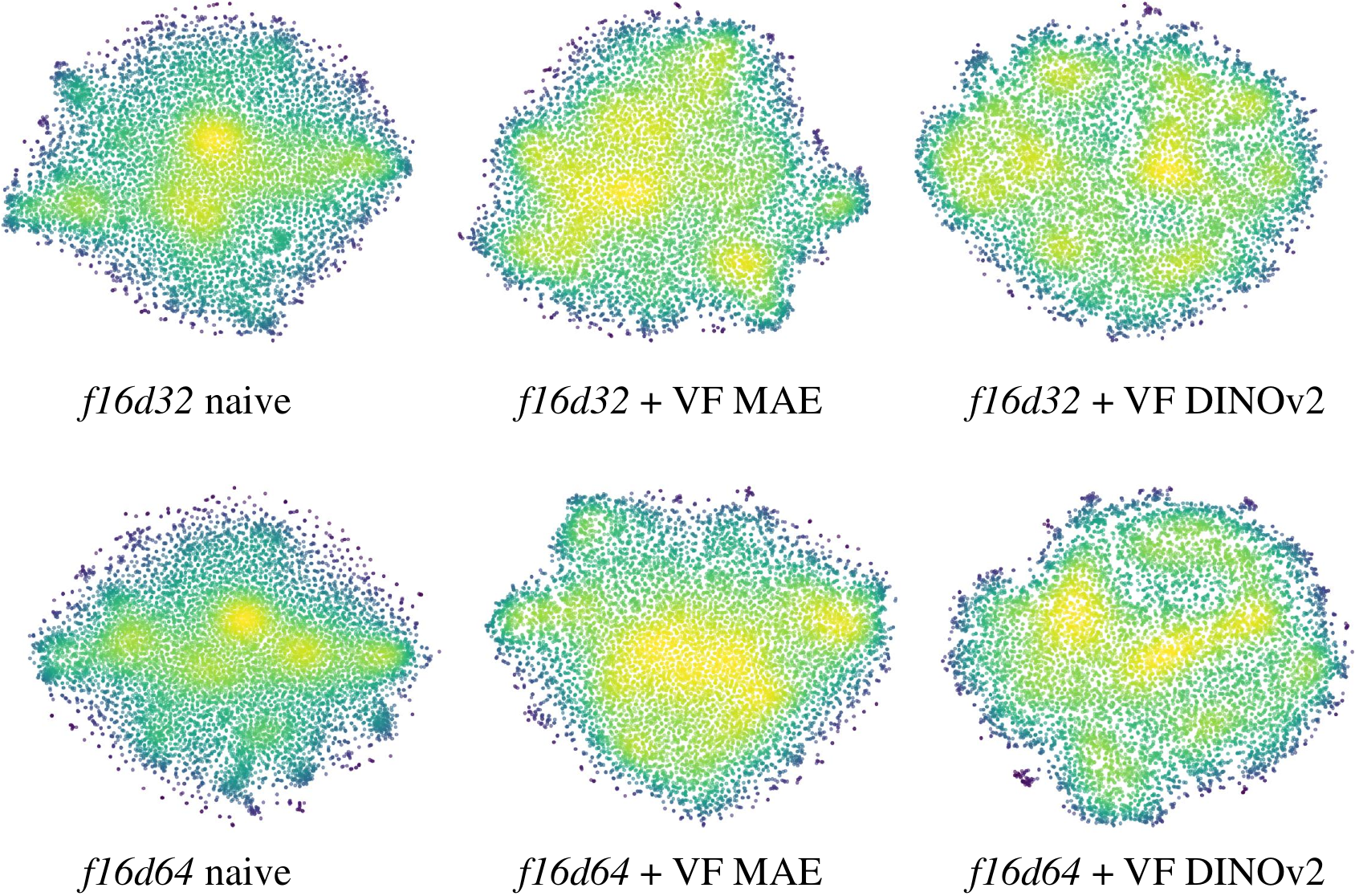}
    \caption{\textbf{Visualization of latent space with t-SNE.} VF loss makes the latent space distribution of high-dimensional tokenizers more uniform.}
    \label{fig:tsne}
\end{figure}

We observe a similar phenomenon in continuous tokenizers. Specifically, we use t-SNE~\cite{t_sne} to visualize the distribution of different latent spaces. Figure~\ref{fig:tsne} shows that VF loss effectively improves the uniformity of the distribution. This observation is further supported by calculating the standard deviation and Gini coefficients of data point distribution using kernel density estimation (KDE) in Tabel~\ref{tab:tsne}. The uniformity metric of the tokenizer seems to be positively correlated with the generative gFID. As the uniformity metric improves, the generative performance of the corresponding tokenizer also increases.

\begin{table}[t]
    \centering
    \renewcommand{\arraystretch}{1.1}
    \resizebox{\linewidth}{!}{
    \begin{tabular}{cl|cccc}
    \toprule
    \textbf{Tokenizer} & \textbf{VF Loss} & \makecell{\textbf{density} \\ \textbf{cv}$\downarrow$} & \makecell{\textbf{gini} \\ \textbf{coefficient}$\downarrow$} & \makecell{\textbf{normalized} \\ \textbf{entropy}$\uparrow$} & \makecell{\textbf{gFID} \\ \textbf{(DiT-B)}$\downarrow$} \\
    \midrule
    \multirow{3}{*}{\textit{f16d32}} & \textit{NaN} & 0.263 & 0.145 & 0.995 & 22.62 \\
         & MAE & 0.193 & 0.101 & 0.997 & 19.89 \\
         & DINOv2 & \textbf{0.178} & \textbf{0.096} & \textbf{0.998} & \textbf{15.82} \\
    \midrule
    \multirow{3}{*}{\textit{f16d64}} & \textit{NaN}  & 0.296 & 0.166 & 0.994 & 36.83 \\
         & MAE & 0.256 & 0.143 & 0.995 & \textbf{23.58} \\
         & DINOv2   & \textbf{0.251} & \textbf{0.141} & \textbf{0.996} & 24.00 \\    
    \bottomrule
    \end{tabular}}
    \caption{\textbf{Relationship between uniformity and generative performance}: We evaluate the uniformity of feature distribution. Results indicate a possible positive correlation between the uniformity of feature distribution and generative performance.}
    \label{tab:tsne}
\end{table}

\section{Conclusion}

This paper focuses on the optimization dilemma in latent diffusion systems. To address the problem, we propose \thevae{}, a VAE aligned with vision foundation models, and \thedit{}, an optimized DiT incorporating advanced design strategies. In \thevae{}, the VF loss function—comprising marginal cosine similarity and distance matrix losses—aligns the VAE's latent space with the vision model, resulting in a more uniform feature distribution and up to 2.8$\times$ faster convergence. 
With \thedit{}, we integrate advanced training techniques and architectural improvements to achieve faster DiT convergence. Combining the high-reconstruction capability of \thevae{} (rFID=0.28) with the rapid convergence of \thedit{}, our approach achieves a state-of-the-art FID of 1.35 on ImageNet 256. Besides, our method achieves 2.11 FID with only 64 epochs, demonstrating 21.8$\times$ speedup to original DiT. To the best of our knowledge, it is the first time that a latent diffusion system could achieve superior reconstruction and generation performance without additional training costs. 
We hope our work could help following research on latent diffusion systems.

{
    \small
    \bibliographystyle{ieeenat_fullname}
    \bibliography{main}
}

\end{document}